\documentclass[a4paper]{llncs}

\usepackage[utf8]{inputenc}

\usepackage{pdflscape}
\usepackage{booktabs}
\usepackage{amsmath}
\usepackage{mathtools}
\usepackage{multirow}
\usepackage[final]{pdfpages}
\usepackage{algorithm}
\usepackage[noend]{algpseudocode}
\floatname{algorithm}{Procedure}

\usepackage{caption}
\usepackage{subcaption}
\usepackage{adjustbox}
\usepackage{amsfonts}
\usepackage{wrapfig}
\usepackage{array}
\usepackage{tabularx}

\usepackage{geometry}
\usepackage{algorithm}
\usepackage{algpseudocode}

\usepackage[english]{babel}

\DeclareMathOperator*{\argmin}{arg\,min}
\DeclareMathOperator*{\argmax}{arg\,max}

\usepackage[%
rm={oldstyle=false,proportional=true},%
sf={oldstyle=false,proportional=true},%
tt={oldstyle=false,proportional=true,variable=true},%
qt=false%
]{cfr-lm}
\usepackage{graphicx}

\usepackage{cite}

\usepackage[T1]{fontenc}

\usepackage[math]{blindtext}

\usepackage{csquotes}

\usepackage{microtype}

\usepackage{url}
\makeatletter
\g@addto@macro{\UrlBreaks}{\UrlOrds}
\makeatother

\usepackage{xcolor}

\usepackage[
bookmarks=false,
breaklinks=true,
colorlinks=true,
linkcolor=black,
citecolor=black,
urlcolor=black,
pdfpagelayout=SinglePage,
pdfstartview=Fit
]{hyperref}
\usepackage[all]{hypcap}

\usepackage{pdfcomment}

\usepackage[capitalise,nameinlink]{cleveref}
\crefname{section}{Sect.}{Sect.}
\Crefname{section}{Section}{Sections}

\usepackage{xspace}

\DeclareFontFamily{U}{MnSymbolC}{}
\DeclareSymbolFont{MnSyC}{U}{MnSymbolC}{m}{n}
\DeclareFontShape{U}{MnSymbolC}{m}{n}{
    <-6>  MnSymbolC5
   <6-7>  MnSymbolC6
   <7-8>  MnSymbolC7
   <8-9>  MnSymbolC8
   <9-10> MnSymbolC9
  <10-12> MnSymbolC10
  <12->   MnSymbolC12%
}{}
\DeclareMathSymbol{\powerset}{\mathord}{MnSyC}{180}

\begin{document}

\input glyphtounicode.tex
\pdfgentounicode=1

\title{A Network Perspective on Stratification of Multi-Label Data}

\author{Piotr Szymański \and Tomasz Kajdanowicz}
\institute{Department of Computational Intelligence, Wrocław University of Technology, Wybrzeże Stanisława Wyspiańskiego 27, 50-370 Wrocław, Poland \\
\email{piotr.szymanski@pwr.edu.pl}, \email{tomasz.kajdanowicz@pwr.edu.pl}}

%

\maketitle

\begin{abstract}
In the recent years, we have witnessed the development of multi-label classification methods which utilize the structure of the label space in a divide and conquer approach to improve classification performance and allow large data sets to be classified efficiently. Yet most of the available data sets have been provided in train/test splits that did not account for maintaining a distribution of higher-order relationships between labels among splits or folds. As a result, classification methods are prone to make mistakes in generalization from data that was not stratified properly. 

We present a new approach to stratifying multi-label data for classification purposes based on the iterative stratification approach proposed by Sechidis et. al. in an ECML PKDD 2011 paper. Our method extends the iterative approach to take into account second-order relationships between labels. Obtained results are evaluated using statistical properties of obtained strata as presented by Sechidis. We also propose new statistical measures relevant to second-order quality: label pairs distribution,  the percentage of label pairs without positive evidence in folds and label pair - fold pairs that have no positive evidence for the label pair. We verify the impact of new methods on classification performance of Binary Relevance, Label Powerset and a fast greedy community detection based label space partitioning classifier. Random Forests serve as base classifiers.  We check the variation of the number of communities obtained per fold, and the stability of their modularity score. Second-Order Iterative Stratification is compared to standard k-fold, label set, and iterative stratification.

The proposed approach lowers the variance of classification quality, improves label pair oriented measures and example distribution while maintaining a competitive quality in label-oriented measures. We also witness an increase in stability of network characteristics.
\end{abstract}


\section{Introduction}\label{sec:intro}

In our recent work \cite{DBLP:journals/entropy/SzymanskiKK16} we proposed a data-driven community detection approach to partition the label space for the multi-label classification. The data-driven approach works as follows: we construct a label co-occurrence graph (both weighted and unweighted versions) based on training data and perform community detection to partition the label set. Then, each partition constitutes a label space for separate multi-label classification sub-problems. As a result, we obtain an ensemble of multi-label classifiers that jointly covers the whole label space. We consider a variety of approaches:  modularity-maximizing techniques approximated by fast greedy and leading eigenvector methods, infomap, walktrap and label propagation algorithms. For comparison purposes we evaluate the binary relevance (BR) and label powerset (LP) - which we call a priori methods, as they a priori assume a total partitioning of the label space into singletons (BR) and lack of any partitioning (LP).

As expected in a paper introducing a new method - we have followed the train/test division set by the providers of benchmarking data sets. Upon closer inspection however we noticed that partitions generated by community detection and the underlying label co-occurrence graphs differed between the train and test data subsets. As many of these data sets were created before first papers on label space partitioning for multi-label classification were published, we find it only natural that data set authors would not take the label structure into account when performing divisions and were providing the best divisions possible with tools available at the time.

Distributing data into train and test sets, or into $k$-fold cross validation is an important element of every experiment. However the method of distribution impacts, and sometimes limits, the generalization ability of evaluated models. The most basic and traditional approach of dividing a data set into $k$ subsets is to define a window of size $m=\frac{n_{samples}}{k}$ and then move the window across the data set - the samples withing the window become the test set in a given fold, while the rest becomes the training set. The window encompasses the first $m$ samples in the data set as in the first fold, the second $m$ samples in the second and so forth. Sometimes random shuffling is employed before performing this method of splitting the data set.

In multi-class classification the above approach may construct divisions that do not provide enough evidence for every class in every fold, due to class imbalance problems, causing artificial bias and limiting the generalization abilities of a model trained on a fold that did not include classes available in the test set. A different approach - called stratification - can be employed to handle this problem. A stratification method divides samples from the data set into roughly equal subsets (strata) that maintain a similar proportion of examples per class over folds.

Ron Kohavi \cite{kohavi1995study} showed that in case of model selection stratification provided better stability than traditional cross-validation methods. Multi-label classification experiments and data set preparations are usually concerned with the problem of dividing available train data into subsets to allow measuring the generalization quality of a model. The \textit{holdout} approach distributes the data into a train, test and (optionally) validation subsets, while the \textit{cross-validation} approach divides data into subsets of equal sizes. Multi-label papers usually follow the original train/test division of provided datasets, in fact a large number of papers follow the data sets provided by the MULAN repository \cite{mulan}. 

Before the work of Sechidis et. al. \cite{sechidis2011stratification} no stratification methods for multi-label data were available and multi-label classification usually followed predetermined train/test splits set by dataset providers, without the analysis in terms of how well the train/test split was performed. As the authors note, lack of control over how multi-label training examples are split into subsets can cause a \textit{lack of positive evidence for a rare label in folds which causes calculation problems in a number of multi-label evaluation measures}. The paper introduces an iterative stratification method that distributes samples based on how desirable a given label is in each fold, tackling the problem of lack of rare label evidence in folds.

The Iterative Stratification (IS) approach is important as folds failing to provide positive samples for rare labels seriously limit models' generalization abilities and does reduce variation of classification quality when using the Binary Relevance approach. The same paper however notes that the stratified yields less variance when Calibrated Label Ranking is used - a method that takes second-order label relations into account.

In this paper we propose an extended version of Iterative Stratification approach, which we call the Second-Order Iterative Stratification, which takes the desirability of label pairs and not just single labels into account when performing stratification. We compare it to IS, stratified and traditional $k$-fold approaches. 

This paper is organized as follows: we describe the proposed stratification method in Section \ref{sec:method} and provide some insight into the second-order label relationship graph based classification methods used for evaluation in Section \ref{sec:classification}. We describe the experimental setup in Section \ref{sec:setup} and describe obtained results in Section \ref{sec:results}. The results section consists of three parts: statistical properties of obtained folds \ref{sec:folds}, stability of obtained label space division using network-based community detection methods \ref{sec:clusters} and variance of classification quality \ref{sec:quality}. We conclude our findings in Section \ref{sec:conclusions}.

\section{Proposed Method}
\label{sec:method}
\begin{algorithm}
	\caption{Second Order Iterative Stratification (SOIS)}
	\label{alg:sois}
	\begin{algorithmic}[1]
		\Procedure{SOIS}{$D, L, k, r$}
        \State $\Lambda = \{\{\lambda_i, \lambda_j\}: (\exists (x, Y) \in D)(\{\lambda_i, \lambda_j\} \subset Y)\}$ \Comment{Incl. i=j, each pair i,j present only once}
		\ForAll{$e \in \Lambda$}
		\State $D^e \gets \{ (x, Y) : Y \cap e \neq \emptyset \}$
		\EndFor
		\For{j=1..k}
		    \State $ c_j \gets |D|*r_j$
		    \ForAll{$e \in \Lambda$}
        		\State $c^{e}_{j}  \gets |D^{e}|*r_j$
    		\EndFor
		\EndFor
        \State \textbf{Return} DistributeOverFolds($D, \Lambda, c$)
		\EndProcedure
	\end{algorithmic}
\end{algorithm}

We propose an extended version of the IS algorithm from Sechidis et. al. that takes into account the second-order label relations in two variants: second-order relations have higher priority in SOIS and the same priority as first-order desirability in SOIS-E. 

In order to introduce the algorithm we start with the following notations. Let $X$ denote the input space, $L$ - the set of labels, $D \subset X \times 2^L$ - the data set, $k$ - the number of desired folds, and $r_{i}{|}_{1}^{k}$ the desired proportion of labels in each of the folds ($\sum_{i=1}^{k}{r_i}=1$). In a typical 10-fold CV scenario: $k=10$, and $r_i=\frac{1}{10}$. Let $E$ denote the set of all pairs of labels that occur together in $D$:

$$ E = \big\{ \{\lambda_i, \lambda_j\}: \big(\exists(\bar{x}, \Lambda) \in D\big) \big(\lambda_i \in \Lambda \wedge \lambda_j \in \Lambda)\big\} $$

The proposed algorithm - Second Order Iterative Stratification (SOIS) first calculates the desired number of samples for each label pair, per fold. Next iterating over label pairs from $E$, it selects the label pair with the least samples available, iterates over all samples with this label pair assigned, assigning the sample to the fold that desires the label pair the most, randomly breaking the ties. The relevant counters of label pair availability and per fold sample, label and label pair desirability are updated, and the internal loop over samples progresses, once all samples evidencing the selected label pair are used up it continues with another iteration of the outer loop. 

Once all label pairs are distributed, the same algorithm is employed to distribute labels from $L$ in a similar manner - which is the graceful fallback to the IS algorithm once all the label pair evidence has been distributed. Once all positive evidence of labels is distributed, negative evidence is randomly distributed as to satisfy sample desirability in each of the folds. SOIS includes both label pairs and labels (represented as ${i, i}$ pairs) in $E$ for consideration. Negative evidence is distributed as in SOIS once all the positive evidence has been distributed.

\begin{algorithm}
	\caption{Iterative distribution of samples into folds}
	\label{alg:idosif}
	\begin{algorithmic}[1]
		\Procedure{DistributeOverFolds}{$D, \Lambda, c$}
		\While {$|\{(x,Y) \in D: Y \neq \emptyset\}|>0$} \Comment{Iterate over positive evidence}
		\ForAll{$\lambda_i \in \Lambda$}
		\State $D^i \gets \{ (x, Y) : Y \cap \lambda_i \neq \emptyset \}$ \Comment{Calculate available samples per label set}
		\EndFor

		\State $l \gets \argmin_{i, D^{i}\neq \emptyset} {|D^i|}$ \Comment{Select the most desirable label set}
		\ForAll{$(x,Y) \in D^l$}
		\State $M \gets \argmax_{j=1\dots |L|}{c^{l}_{j}}$ \Comment{Select most desirable fold based on label set desirability}
		\If {$|M|==0$}
		\State $m \gets onlyElement(M)$
		\Else
	    \State $M' \gets \argmax_{j \in M}{c_{j}}$  \Comment{Split the tie by selecting the fold that need new samples most}
	    \If {$|M'|==0$}
	        \State $m \gets onlyElementOf(M')$
	    \Else
	        \State $m \gets randomElementOf(M')$ \Comment{Split the tie randomly}
	    \EndIf
		\EndIf
		\State $S_m \gets S_m \cup (x, Y)$ \Comment{Update sets}
		\State $D \gets D \setminus (x, Y)$
		\State $c^{l}_m \gets c^{l}_m - 1$ \Comment{Update desired number of examples in c}
		\State $c_m \gets c_m - 1$
		\EndFor
		\EndWhile
		\ForAll{$(x,Y) \in D$} \Comment{Distribute negative examples}
		\State $M \gets \argmax_{i=1\dots k}{c_{i}}$ \Comment{Select fold that desires most samples}
		\State $m \gets randomElementOf(M)$ \Comment{Split tie randomly if needed}
		\State $S_m \gets S_m \cup (x, Y)$ \Comment{Update sets}
		\State $c_m \gets c_m - 1$ \Comment{Update desired number of examples in c}
		\EndFor
        \State \textbf{Return} $S_1, \dots, S_k$
		\EndProcedure
	\end{algorithmic}
\end{algorithm}

\section{Classification Methods}
\label{sec:classification}

In our experimental setting we compare the generalization stability of stratification approaches using Binary Relevance, Label Powerset and two variants of our community-detection based label space partitioning Label Powerset scheme. As the first two classifiers are obvious, we provide an description of our method.

The training phase starts with constructing the label co-occurrence graph as follows. We start with and undirected co-occurrence graph $G$ with the label set $L$ as the vertex set and the set of label pairs $E$ is the set of edges. In the weighted variant, we extend this unweighted graph $G$ to a weighted graph by defining a weight function $w: L \rightarrow \mathbb{N}$ that counts the number of samples with positive evidence for a pair of labels in the training set:
$$ w(\lambda_{i},\lambda_{j}) =\Big|  \big\{ \bar{x}: (\bar{x}, \Lambda) \in D \wedge \lambda_i \in \Lambda \wedge \lambda_j \in \Lambda \wedge i\neq j \big\}\Big|$$

For such graph a $G$, weighted or unweighted, we find a label space division by using a community detection method that partitions the graph's vertex set which is equal to the label set. In this paper we use the fast greedy modularity-maximization approach \cite{clauset_finding_2004}. It detects a partition of $L$ that tries to greedily maximize the modularity measure by \cite{newman_finding_2003}. Behind this measure lies an assumption that \textit{true community structure in a network corresponds to a statistically surprising arrangement of edges} \cite{newman_finding_2003}, i.e. that a community structure in a real phenomenon should exhibit a structure different than an average case of a random graph, which is generated under a given \textit{null model}. A well-established null model is the configuration model, which joins vertices together at random, but maintains the degree distribution of vertices in the graph. 

For a given partition of the label set, the modularity measure is the difference between how many edges of the empirically observed graph have both ends inside of a given community, i.e. $e(C) = \{(u,v) \in E: u \in C \wedge v \in C \}$ versus how many edges starting in this community would end in a different one in the random case: $r(C) = \frac{\sum_{v \in C}{deg(v)}}{|E|} $. More formally this is $Q(C) = \sum_{c \in C} {e(c) - r(c)}$. In case of weights, instead of counting the number of edges the total weight of edges is used and instead of taking vertex degrees in $r$, the vertex strenghts are used - a precise description of weighted modularity can be found in Newman's paper \cite{newman2004analysis}.

The fast greedy approach starts with each node as a singleton community and merges the communities iteratively. In each iteration the algorithm merges two communities based on which merge achieves the highest contribution to modularity. The algorithm stops when there is no possible merge that would increase the value of the current partition's modularity.

All in all, in our data-driven scheme, the training phase is performed as follows:
\begin{enumerate}
\item the label co-occurence graph is constructed based on the training data set
\item the selected community detection algorithm is executed on the label co-occurence graph
\item for every community $L_i$ a new training data set $D_i$ is created by taking the original input space with only the label columns that are present in $L_i$
\item for every community a classifier $h_i$ is learned on training set $D_i$
\end{enumerate}

The classification phase is performed by performing classification on all subspaces detected in the training phase and taking the union of assigned labels: $h(\bar{x}) = \bigcup_{j=0}^{k} { h_j(\bar{x})}$.

\section{Experimental Setup}
\label{sec:setup}
We perform experimental evaluation of presented stratification approaches on 16 benchmark data sets that were available in the MULAN repository \cite{mulandata}: Corel5k, bibtex, delicious (not used in network approaches as calculations did not finish), emotions, enron, genbase, mediamill, medical, rcv1subset1, rcv1subset2, rcv1subset3, rcv1subset4, rcv1subset5, scene, tmc2007-500, yeast. Experiments were performed using the scikit-multilearn library by \cite{scikitmultilearn}.

Stratification methods were evaluated by analyzing the characteristics of each fold in terms of statistical measures, classification quality measures with classification performed using Binary Relevance, Label Powerset and Data-Driven Label Space Partitioning with Label Powerset.

Additionaly we evaluate the impact of stratification methods on a models' ability to perform generalization approaches using classification and label ranking quality metrics provided by the scikit-learn library by \cite{scikit-learn}. Evaluated models include Binary Relevance, Label Powerset, and data-driven label space partitioning following our previous research.

\section{Results}
\label{sec:results}
We evaluate the considered stratification methods in terms of three types of properties. First we are interested in the quality of sample distribution over folds in terms of the statistical properties of output spaces that the model will work on in a cross-validation setting. Next we evaluate what is the impact of generalization quality in two baseline approaches - Binary Relevance which should depend on how well each of the label is evidenced and counterevidenced in each fold, and Label Powerset which should be more prone to higher-order relation misstratification. Finally we look into the stratification methods' impact on label co-occurrence graphs, the detected communities, their stability, the obtained modularities and generalization quality of under the partitioned scheme.

\subsection{Statistical properties of folds}
\label{sec:folds}
In this section we compare sampling approaches using statistical properties of obtained data subsets using the properties from Sechidis et. al.'s paper and also their second-order label relations equivalents. We follow the notation from previous paragraphs to define the measures used in this section.

Label Distribution (LD) is a measure that evaluates how the proportion of positive evidence for a label to the negative evidence for a label deviates from the same proportion in the entire data set, averaged over all folds and labels. In the following notation $S^{i}_{j}$ and $D^{i}$ are the sets of samples that have the $i$-th label from $L$ assigned in the $j$-th fold and the entire data set, respectively:

$$ LD = \frac{1}{|L|}\sum_{i=1}^{|L|}{\Big(\frac{1}{k}\sum_{j=1}^{k}{\Big|\frac{|S^{i}_{j}|}{|S_j|-|S^{i}_{j}|}-\frac{|D^{i}|}{|D|-|D^{i}|}\Big|}\Big)} $$

Label Pair Distribution (LPD) is an extension of the LD measure that operates on positive and negative subsets of label pairs instead of labels. In the following definition $S^{i}_{j}$ and $D^{i}$ are the sets of samples that have the $i$-th label pair from $E$ assigned in the $j$-th fold and the entire data set, respectively:

$$ LPD = \frac{1}{|E|}\sum_{i=1}^{|E|}{\Big(\frac{1}{k}\sum_{j=1}^{k}{\Big|\frac{|S^{i}_{j}|}{|S_j|-|S^{i}_{j}|}-\frac{|D^{i}|}{|D|-|D^{i}|}\Big|}\Big)} $$

Examples Distribution (ED) is a measure of how much a given fold's size deviates from the desired number of samples in each of the folds:

$$ ED = \frac{1}{k}\sum_{j=1}^{k}{\Big||S_j|-c_{j}\Big|} $$

In a cross-validation setting we are also interested in how, we thus define:
\begin{itemize}
    \item FZ - the number of folds that contain at least one label with no positive examples
    \item FLZ - the number of fold-label pairs with no positive examples
    \item FLPZ - a second-order extension of FLZ - the number of fold - label pair pairs with no positive examples
\end{itemize}
In the case of FLPZ, as it happens that label pairs do not have enough evidence to split over the evaluated 10 folds, we only count these label pair - fold pairs that had more folds without positive examples, than the inevitable minimum value corresponding to the number of folds minus the number of available samples with a label pair.

\begin{table}[h]
 \centering
 \begin{tabular}{lrrrrrrrrrr}
\toprule
{} &  kfold &       & labelset &       & SOIS &       &       IS &       \\
{} &   mean &   std &       mean &   std &   mean &   std &      mean &   std \\
\midrule
Corel5k     &  0.828 &  0.04 &      0.820 &  0.28 &                \textbf{0.699} &  \underline{0.01} &     0.709 &  \underline{0.01} \\
bibtex      &  0.694 &  0.03 &      0.851 &  0.29 &                \textbf{0.662} &  \underline{0.02} &     0.687 &  \underline{0.02} \\
delicious   &  0.592 &  \underline{0.00} &      0.887 &  0.30 &                \textbf{0.582} &  \underline{0.00} &     0.584 &  \underline{0.00} \\
emotions    &  0.285 &  0.11 &      0.256 &  0.14 &                \textbf{0.161} &  \underline{0.04} &     0.251 &  0.09 \\
enron       &  0.649 &  0.07 &      0.806 &  0.28 &                \textbf{0.578} &  \underline{0.02} &     0.602 &  \underline{0.02} \\
genbase     &  0.686 &  0.15 &      0.601 &  0.31 &                \textbf{0.487} &  0.16 &              0.494 &  \underline{0.14} \\
mediamill   &  0.491 &  0.03 &      0.596 &  0.23 &                \textbf{0.324} &  \underline{0.01}  &     0.364 & \underline{0.01} \\
medical     &  0.762 &  0.06 &      0.762 &  0.30 &                \textbf{0.736} &  \underline{0.03} &     0.751 &  0.04 \\
rcv1subset1 &  0.712 &  0.02 &      0.729 &  0.26 &                \textbf{0.581} &  \underline{0.01} &     0.606 &  0.02 \\
rcv1subset2 &  0.712 &  0.05 &      0.727 &  0.26 &                \textbf{0.574} &  \underline{0.01} &     0.598 &  0.02 \\
rcv1subset3 &  0.721 &  0.04 &      0.731 &  0.26 &                \textbf{0.583} &  \underline{0.01} &     0.606 &  0.02 \\
rcv1subset4 &  0.720 &  0.08 &      0.709 &  0.26 &                \textbf{0.574} &  \underline{0.01} &     0.600 &  0.02 \\
rcv1subset5 &  0.714 &  0.03 &      0.732 &  0.26 &                \textbf{0.584} &  \underline{0.02} &     0.603 &  \underline{0.02} \\
scene       &  0.711 &  0.10 &      0.277 &  0.11 &                \textbf{0.276} &  \underline{0.05} &     0.312 &  0.14 \\
tmc2007-500 &  0.218 &  0.02 &      0.347 &  0.17 &                \textbf{0.159 }&  \underline{0.01} &     0.207 &  0.03 \\
yeast       &  0.078 &  0.03 &      0.095 &  0.04 &                \textbf{0.062} &  \underline{0.01} &     0.064 &  0.02 \\
\bottomrule
\end{tabular}

\caption{Percentage of label pairs without positive evidence, averaged over 10 folds, with standard deviation. The lesser the better. The best performing division method in bold. Methods with smallest variance are underlined.\label{tab:stats.percentages}}
\end{table} 

As there is little reason to generalizing FZ to label pairs as an integer measure, because all folds miss at least one label pair, we generalize it as a measure of percentage of label pairs that are not present in each of the folds. We provide average percentages per data set per method alongside with standard deviations in Table 
\ref{tab:stats.percentages}. 

\begin{figure}[h!]
\includegraphics[width=1\textwidth]{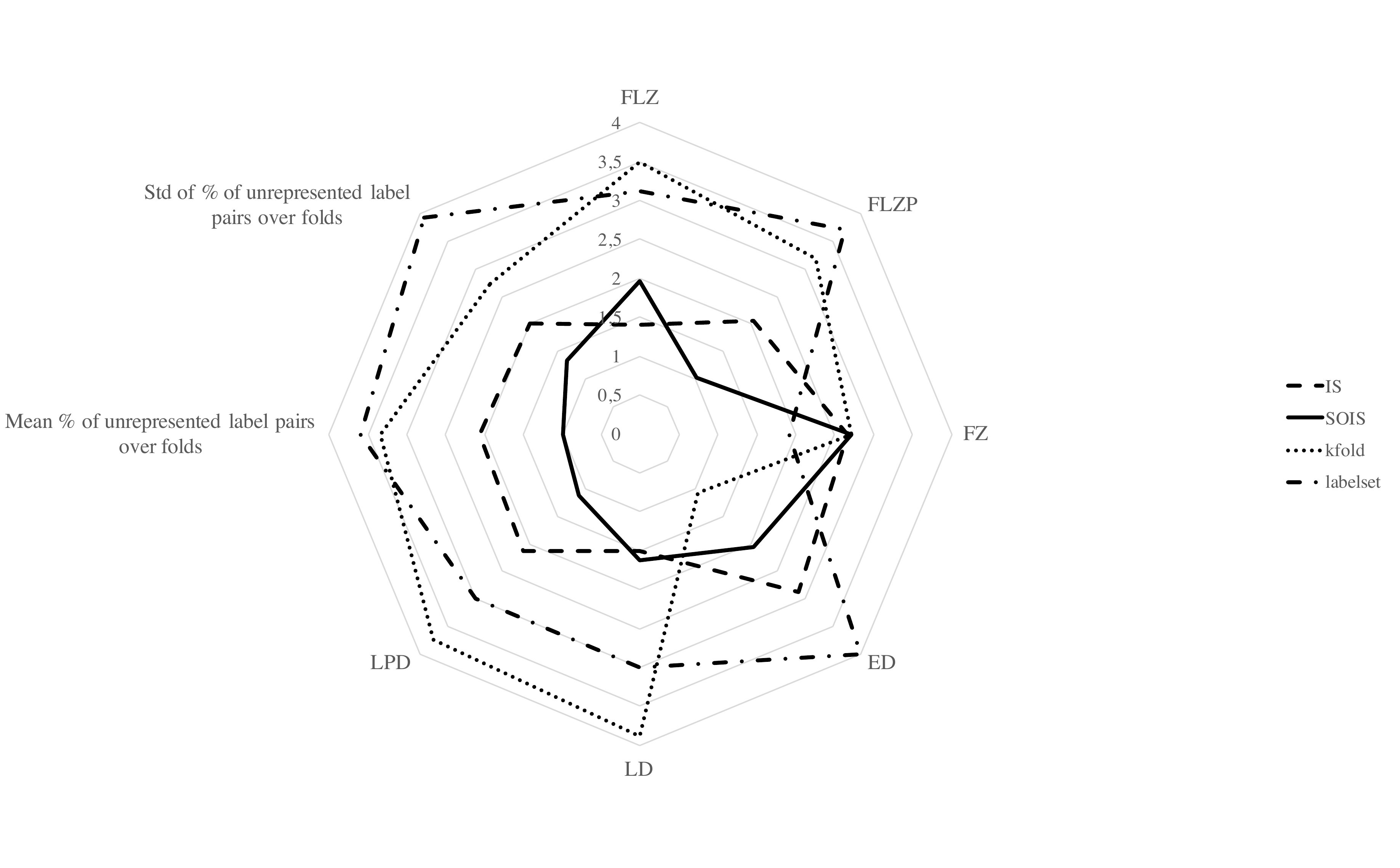}
\centering
\caption{Average ranks of proposed stratification approaches with regard to statistical properties of generated strata.\label{fig:distributions}}
\end{figure}

The best method for multi-label stratification should provide folds that have a small Example, Label and Label Pair Distribution scores, as such a stratification remains well balanced both in terms of evidence and in terms of size. It should also yield small number of folds that miss evidence for labels and label pairs and preferably if a miss happens it should be as small as possible, thus FZ, FLZ and FLZP should be as small as possible. Similarly the percentage of label pairs not evidenced per fold should be both small on average, but also stable. Let us look at Figure \ref{fig:distributions} to see how the evaluated methods rank on average from the statistical properties perspective.

The k-fold approach is a clear winner when it comes to lowest deviation fold sizes (ED) which does not surprise us, as the only criterion of the traditional k-fold division is the number of examples. While simplest, available in practically all multi-label classification libraries and thus most often used - it remains the worst ranked in FLZ, LD, LPD. It also ranks second worst in terms of FLZP and the percentage of label pairs not evidenced per fold both on average and in the scale of standard deviation of the percentage of label pairs not evidenced per fold. It is noteworthy that in most of the evaluated data sets this approach generates folds in which, on average, lack positive samples for 70-80\% of label pairs. k-fold does not provide folds that maintain a distribution of labels or label pairs. Clearly this measure should only be used when the data set authors have taken other precautions concerning label and label pair distributions before performing division. 

The stratified labelset approach ranks on par with the best in FZ, worst in ED, FLZP and coverage of label percentages with positive evidence - both in average and standard deviation. It ranks second worse in other measures. In practice what we have observed is that there this approach creates the most informed fold first. That fold contains positive evidence for as many label combinations (classes) as possible, leaving few samples to serve as such evidence in other folds. Such an approach yields large deviation of percentages of unevidenced label pairs and also creates a disproportion in fold sizes. It is succesful in minimizing FZ as the first fold is always sure to be well-evidenced. This method should only be used in the case when there is little to no imbalance of positive evidence distribution among labelsets, in practice - never.

The Iterative Stratification approach ranks best in terms of FLZ and LD, and on par with the best in terms of FZ. It performs second best in label pair measures, but it ranked visibly worse than SOIS. This approach performs best stratification when it comes to making sure that all labels have positive evidence in all folds, but underperforms when it comes to positive evidence for label pairs. It also ranks second worse in ED losing only to stratified labelset approach.

Second-Order versions of IS (SOIS) perform best in measures related to label pairs and is better in ED than other non-kfold approaches, while also performing second best in all other measures, ranking closely to the best performers.

Out of the two methods scoring well in label and label pair measures, SOIS is clearly a better choice as the gain in stability of label pair measures is larger than the loss in FLZ. In other single-label measures SOIS ranks closely to IS. The method successfuly finds a compromise between evidencing labels and label pairs while maintaining small deviations of sample sizes per fold.

\begin{figure}[h!]
\includegraphics[width=1\textwidth]{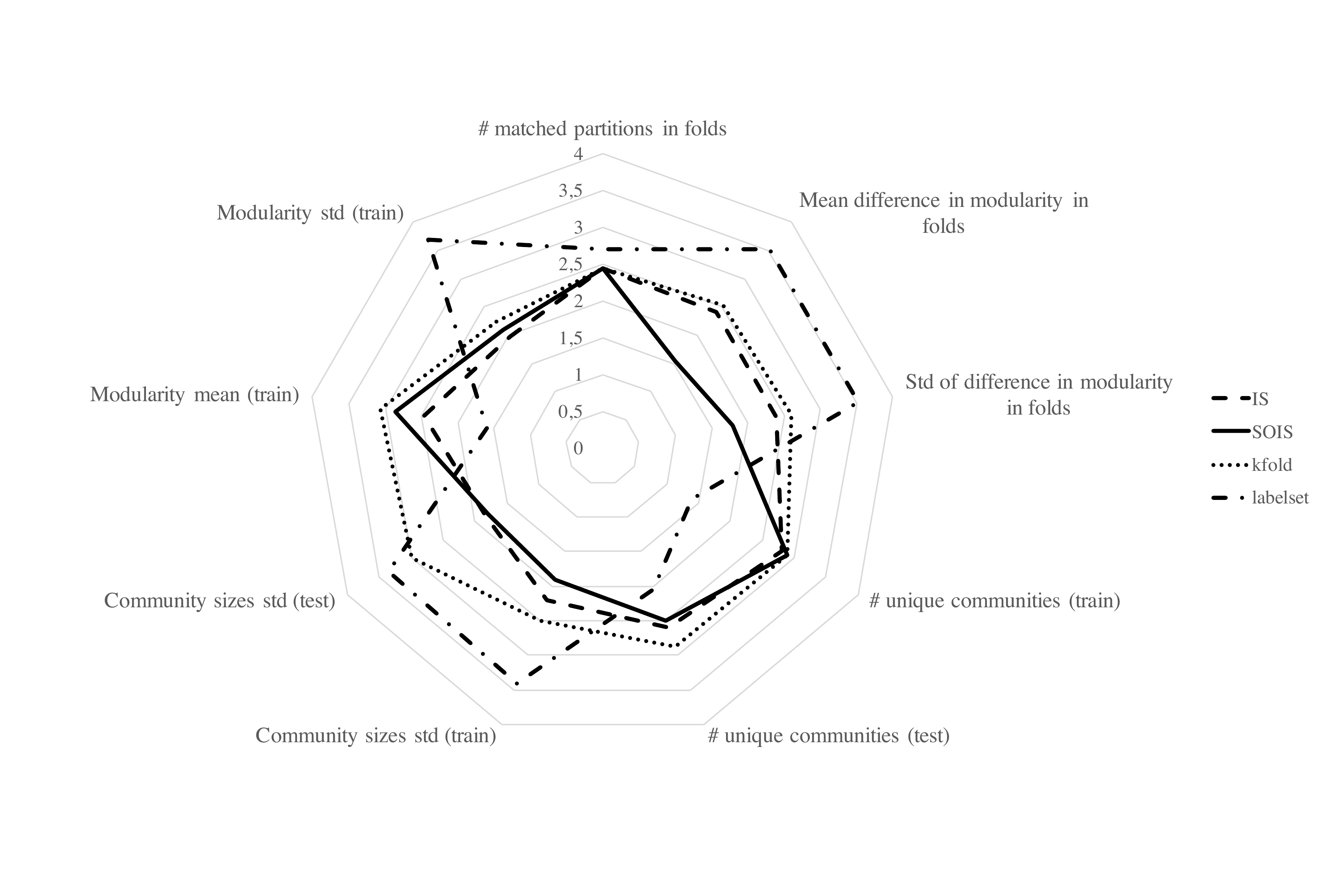}
\centering
\caption{Average ranks of proposed stratification approaches with regard to different  characteristics of the unweighted label co-occurence graph constructed on stratified folds.\label{fig:network.fg}}
\end{figure}

\subsection{Stability of Network Characteristics}
\label{sec:clusters}

From the Network Perspective it is important that a stratification methods provides stability in obtained modularity scores both over the training folds and between train and test folds of a given strata. In the perfect case the stratification algorithm should provide data that allow constructing graphs similar enough that the community detection algorithm would find exactly the same community in all folds, and exactly the same community in every train/test fold pair per stratum.

We used the fast greedy modularity maximization scheme provided by the igraph \cite{igraph} library to detect communities on label graphs constructed from training and test examples in each of the folds. We constructed both the unweighted and weighted graphs and performed community detection on both of them. We review the case of each of the graphs separately. 

\begin{figure}[h!]
\includegraphics[width=1\textwidth]{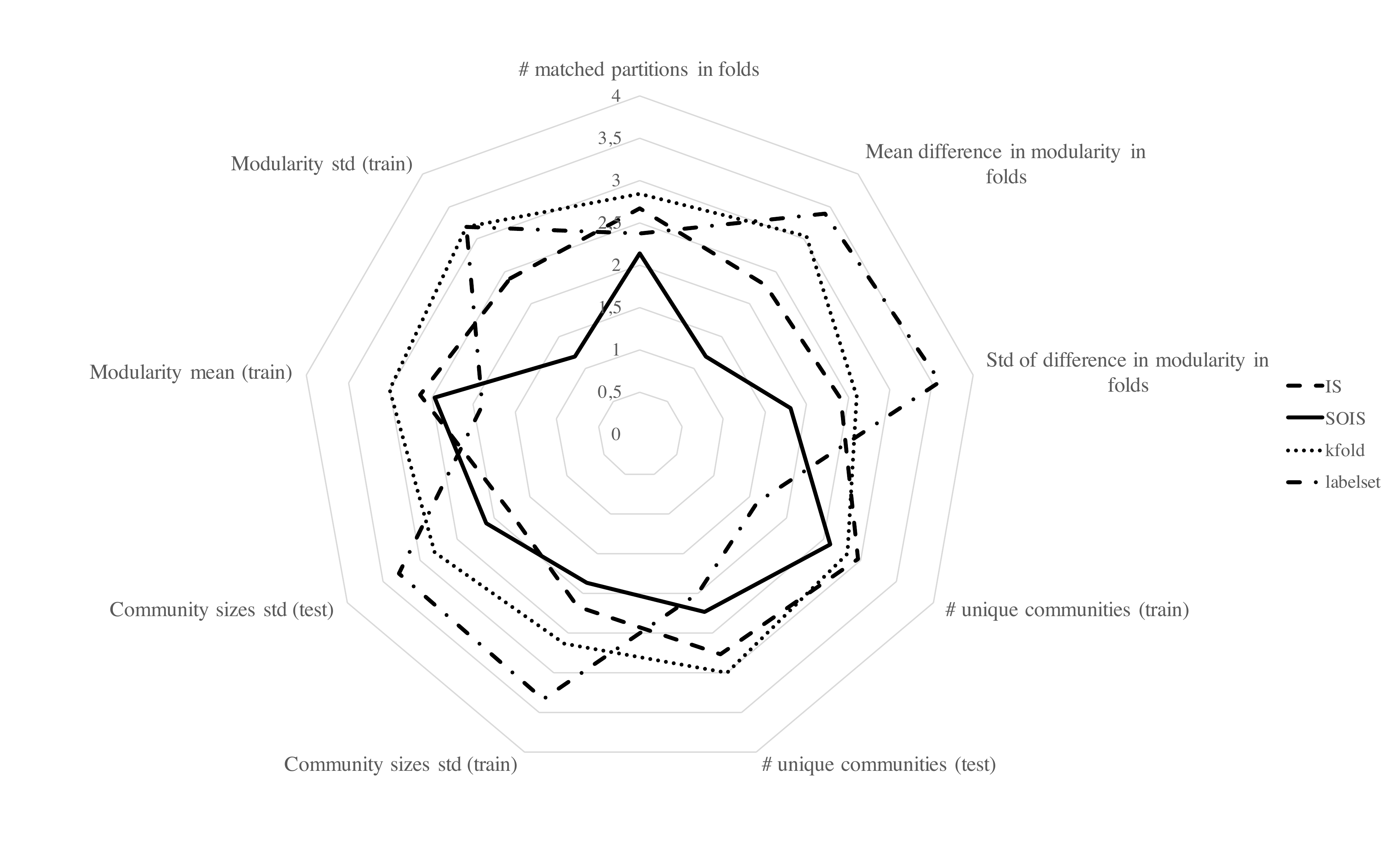}
\centering
\caption{Average ranks of proposed stratification approaches with regard to different  characteristics of the weighted label co-occurence graph constructed on stratified folds.\label{fig:network.fgw}}
\end{figure}

We evaluated the following Network Characteristics: the mean and standard deviation of modularity scores over training folds, the stability (i.e. the standard deviation) of the number of sizes of communities detected in each train and test fold and the number of unique communities. We also count the number of partitions that were exactly matched per train-test subsets of every fold and the mean and standard deviation of modularity differences between train-test subsets of every fold. The results are illustrated in Figure \ref{fig:network.fg} for the unweighted graph case and in Figure \ref{fig:network.fgw} for the weighted case.

The labelset stratification approach ranks best when it comes to obtained modularity mean on train examples, yet worst when it comes to standard deviation of the modularity score. It is like this because the first fold is always provided with as complete evidence as possible, which makes any mean score higher, while other folds do not include rare data and become different problems - yielding a very high standard deviation. Similar case happens with unique communities, where the problems with less evidence become more similar, yet simpler, yielding less communities due to lack of edges - as in this case edges are binary indications of existence of samples labeled with a given label pair. In all other measures the labelset approach performs worst and, as was in the statistical measures case, should not be used in practice.

When labelset stratification is discarded kfold become the consistently worst performing stratification method. It ranks last in all measures, while ranking close to iterative approaches in stability of unique communities and matched partitions it underperforms in modularity differences between train and test subsets and community sizes.

IS and SOIS are on par in most measures, IS shows a slight advantage in the standard deviation while SOIS distances IS more significantly in modularity differences in train-test pairs.

In the weighted variant we observe similar behaviour of the labelset stratification. K-fold ranks worst in all measures apart from the number of unique communities detected, where it is second worst but very close to IS.

SOIS ranks higher than SOIS in every measure apart from the standard deviation of obtained community sizes in test sets and is only slightly higher ranked in terms of mean obtained modularity on train data. We see that SOIS ranks consistently better in better matching of partitions between relevant train-test pairs and yielding lower and more stable modularity differences among these pairs. It also is much more stable when it comes to modularity standard deviation while obtaining a better but similar modularity mean average rank. SOIS yields ranks better in number of unique communities detected, while it does not differ much in community sizes in train and test folds. 

We observe that SOIS is closer to realizing the ideal scenario than IS in most of the measures on weighted graphs, where the number and not just the presence of samples is most important. SOIS also maintains and advantage in terms of unweighted graphs, but the difference with IS is less significant. Similarly as in the case of statistical measures we note that SOIS is a better choice than IS when it comes to maintaining network characteristics across folds.

\subsection{Variance of generalization quality}
\label{sec:quality}

\begin{figure}[h]
\includegraphics[width=1\textwidth]{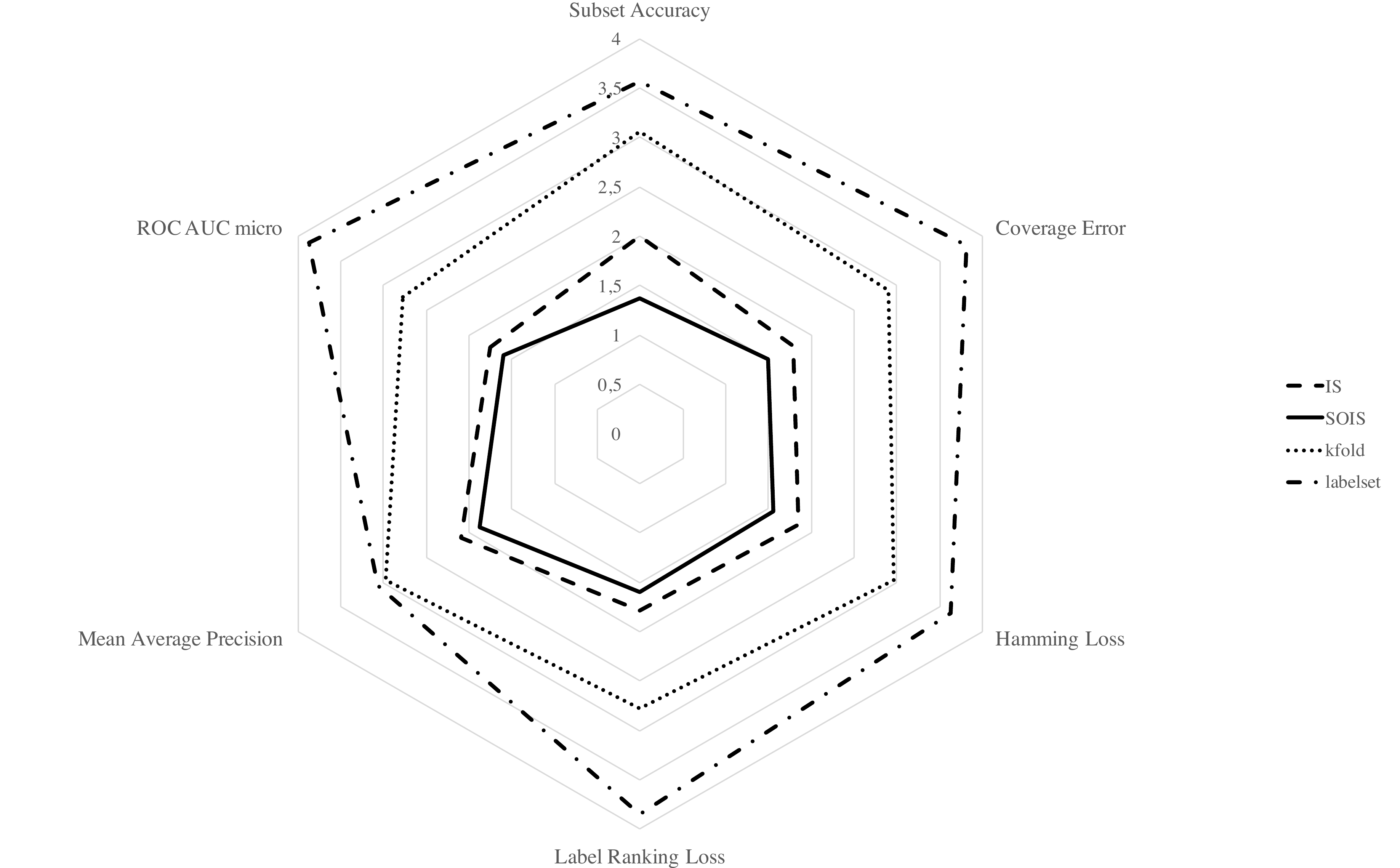}
\centering
\caption{Average ranks of proposed stratification approaches with regard to standard deviation of scores in evaluated generalization measures when classification was performed using Binary Relevance over stratified folds.\label{fig:class.br}}

\includegraphics[width=1\textwidth]{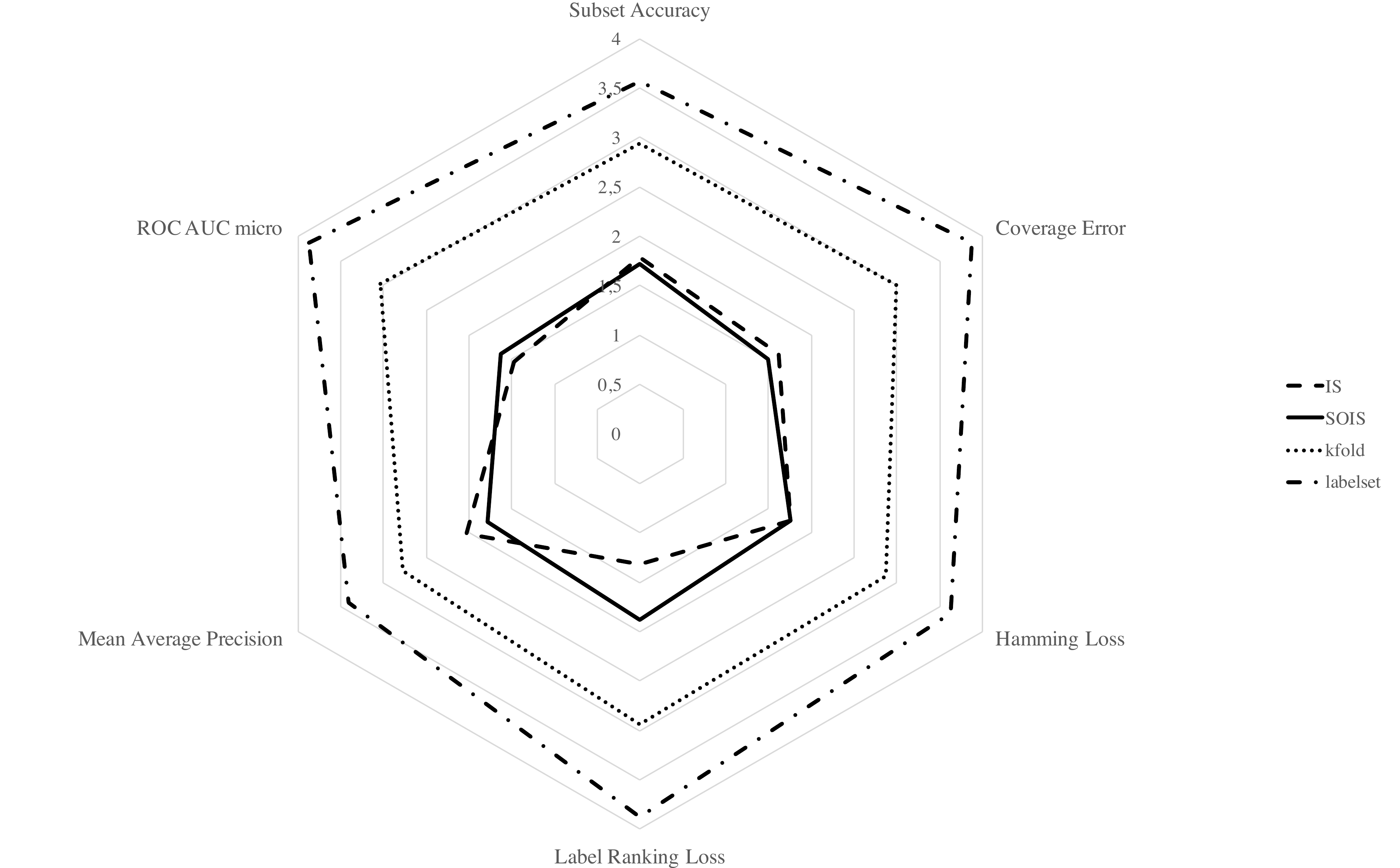}
\centering
\caption{Average ranks of proposed stratification approaches with regard to standard deviation of scores in evaluated generalization measures when classification was performed using Label Powerset over stratified folds.\label{fig:class.lp}}
\end{figure}

\begin{figure}[h]
\includegraphics[width=.9\textwidth]{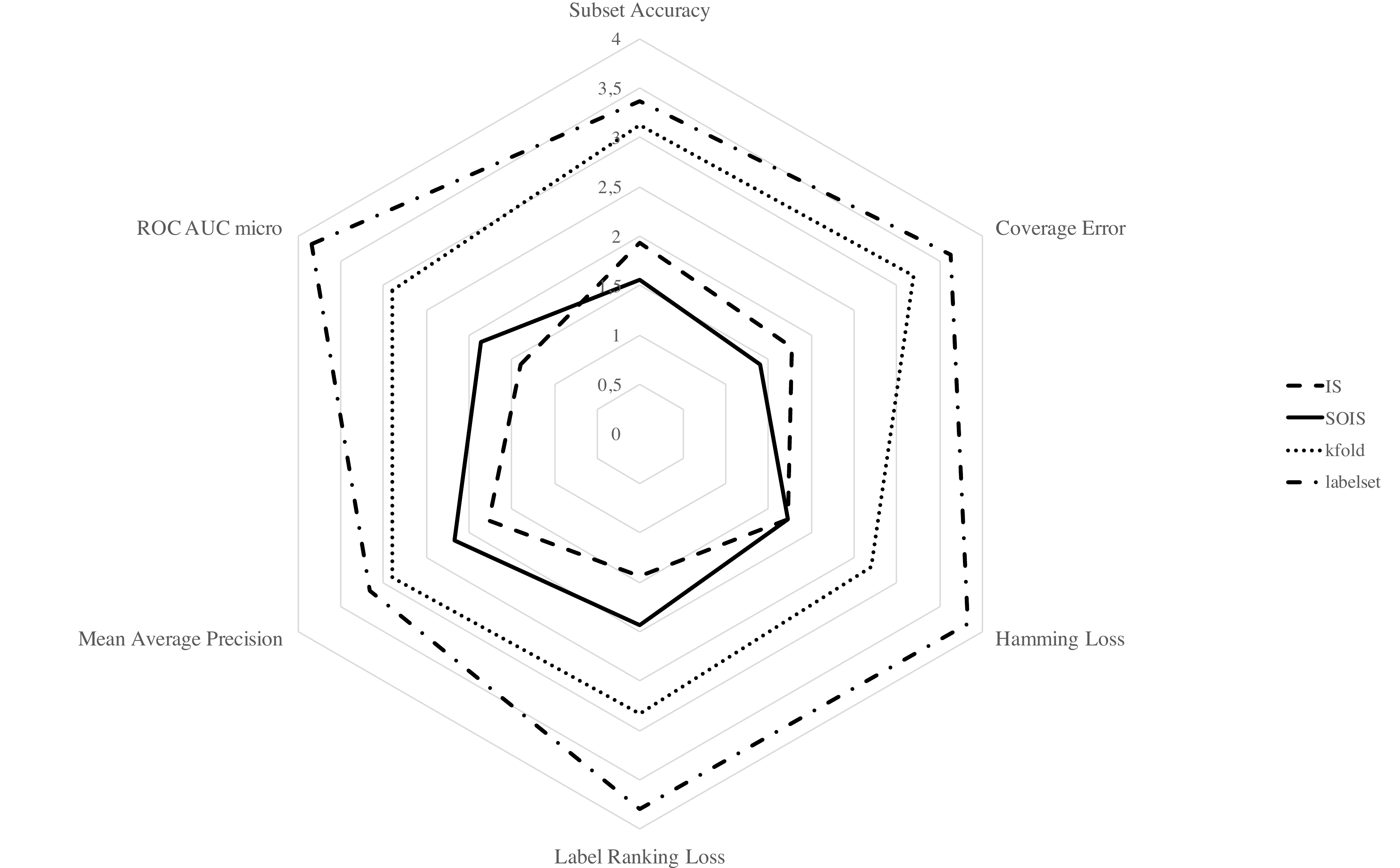}
\centering
\caption{Average ranks of proposed stratification approaches with regard to standard deviation of scores in evaluated generalization measures when classification was performed using fast greedy community detection on unweighted graphs over stratified folds.\label{fig:class.fg}}

\includegraphics[width=.9\textwidth]{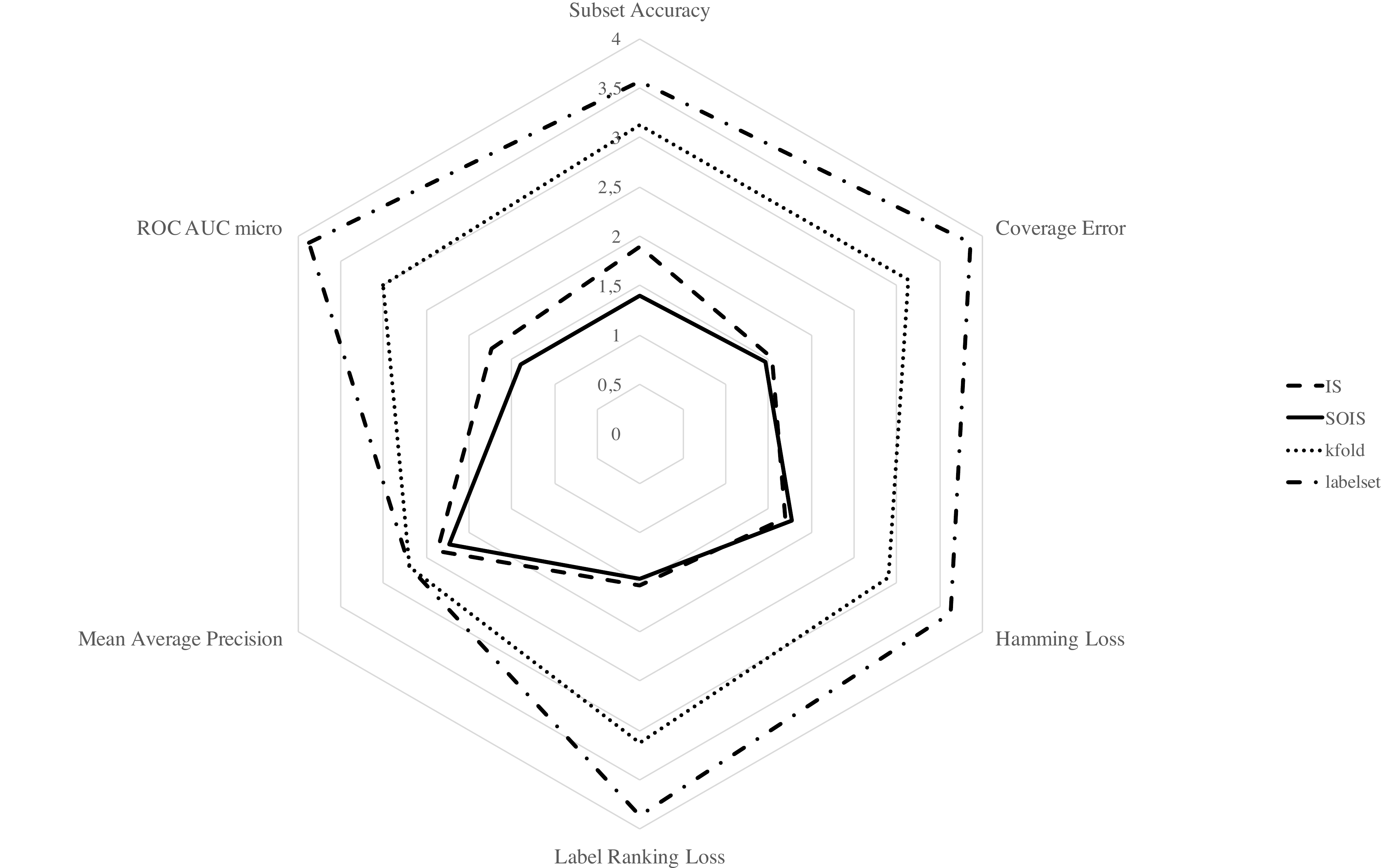}
\centering
\caption{Average ranks of proposed stratification approaches with regard to standard deviation of scores in evaluated generalization measures when classification was performed using fast greedy community detection on weighted graphs over stratified folds.\label{fig:class.fgw}}
\end{figure}

In terms of generalization quality one would expect for stratification methods to allow comparable generalization perspectives to the model in each of the folds, while not compromising the average generalization quality. For evaluation purposes we take two standard approaches to classification - Binary Relevance (BR, Figure \ref{fig:class.br}) and Label Powerset (LP, Figure \ref{fig:class.lp}), and two wariants of the data-driven label space clustering using fast greedy modularity maximization on unweighted (FG, Figure \ref{fig:class.fg}) and weighted (FGW, Figure \ref{fig:class.fgw}) label co-occurrence graphs. We do not compare them to each other, instead we compare the standard deviation of their generalization quality over folds generated by each stratification method. We recall the original measures presented in Sechidis et. al.'s work: Subset Accuracy, Coverage Error, Hamming Loss, Label Ranking Loss, Mean Average Precision (also known as macro-averaged precision), micro-averaged Receiver Operating Characteristic Area Under Curve.

The Binary Relevance case is fairly evident, with label powerset yielding the highest standard deviations, followed by k-fold and IS, while SOIS ranks best with most stable generalization.

The Label Powerset provides similar worst-performing picture of labelset and kfold approaches. In this case however the distances between the best ranked IS and SOIS are small and what SOIS gains in Mean Average Precision or Coverage Error or Subset Accuracy it loses in Label Ranking Loss. From a practical point of view the methods perform equally well in this case.

Similarly to the case of network characteristics when it comes to unweighted fast greedy case, the standard deviation of generalization scores is similar between IS and SOIS, while the other methods rank last and second last. In this case again IS ranks better in Label Ranking Loss, Mean Average Precision and ROC AUC micro, while SOIS ranks better in Subset Accuracy and Coverage Error. The distances between the ranks are small and what one method gains in one measure's stability, it loses in the another one.

In the case of weighted label co-occurence graphs we observe that, consistently with other experimental results, kfold and stratification approaches rank worst when it comes to standard deviation of generalization measures. In this case we also observe, what is compatible with the network characteristics results for the weighted graph, that SOIS ranks better or on par than IS becoming a stratification method of choice.

We observe that kfold and stratification methods perform worst in classification stability across all evaluated cases. The two algorithms that perform best: IS and SOIS provide similar generalization stability with Label Powerset or unweighted fast-greedy scheme. When Binary Relevance or weighted fast-greedy approach are used, SOIS performs better.

We did not provide the result tables in print as all of the data and result tables are available in the Github repository associated with this paper\footnote{\url{https://github.com/niedakh/multilabelstratification}} to maintain the standards of reproducibility. In print the tables would span multiple pages and would not serve the purpose of illustrating the results in an understandable fashion. The result tables, notebooks and code are thus provided in the repository and can be browsed digitally to allow comfortable review.

\section{Conclusions and future research}
\label{sec:conclusions}

Our experiments show that the stratification based on label powerset transformation results in distributing as much positive evidence as available in the first fold(s) and running out of positive evidence for other folds. Thus the data set actually becomes divided into completely different problems - a more complicated one based on the super fold, and the easier in the other folds. While such a division allows better scoring due to lack of hard test samples in most of the folds, it is far from providing data that allow a stable generalization. The other traditional method for data set division - kfold - ranks consistently worst in terms of the scale of standard deviation obtained in evaluated measures.

We discourage the use of k-folding and label powerset transformation based stratification and instead propose to use an iterative approach that takes second-order relationships into account and provides folds that exhibit more stability in terms of statistical measures, network characteristics and generalization quality. We recommend using SOIS instead of IS as the stability increase yielded by SOIS is usually a greater advantage then the rare cases of SOIS ranking lower that IS, noting the small distances in ranks in these cases.

Future research into the topic should examine the impact on other community detection and clustering methods for example $k$-means, infomap, etc, a larger number of data sets and stronger theoretical considerations.

\section*{Acknowledgments}\label{sec:Acknowledgments}
The work was partially supported by The National Science Centre the research projects no. 2016/21/N/ST6/02382 and 2016/21/D/ST6/02948 and by the Faculty of Computer Science and Management, Wrocław University of Science and Technology statutory funds. Research financed under the European Union’s Horizon 2020 research and innovation programme under the Marie Skłodowska-Curie grant agreement No 691152[RENOIR] and the fund for supporting internationally co-financed projects in 2016-2019.  

\bibliographystyle{splncs03}
\bibliography{paper}

\end{document}